\documentclass[runningheads]{llncs}

\usepackage{eccv}

\usepackage{eccvabbrv}

\usepackage{graphicx}
\usepackage{booktabs}

\usepackage[accsupp]{axessibility}  
\usepackage{multirow, multicol}

\usepackage{hyperref}

\usepackage{orcidlink}

\begin{document}

\title{OmniX:$\!$ Any-view$\!$ and$\!$ Any-time 4D Reconstruction$\!\!$ via Feed-forward Trajectory Fields} 

\titlerunning{Omnix: 4D Reconstruction via Trajectory Fields}

\newcommand{\authormark}[1]{\textsuperscript{#1}}
\newcommand{\authorfootnote}[2]{%
  \begingroup
  \renewcommand{\thefootnote}{#1}%
  \footnotetext{#2}%
  \endgroup
}

\author{Yanqin Jiang\inst{1,2}\authormark{\dag} \and Tengfei Wang\inst{3}\authormark{\ddag} \and Zhengwei Wang\inst{3} \and Chenjie Cao\inst{3} \and Junta Wu\inst{3} \and Wenhan Luo\inst{4} \and Weiming Hu\inst{1,2,5,6} \and Jin Gao\inst{1,2,5}\authormark{\ddag} \and Chunchao Guo\inst{3}}

\authorrunning{Y. Jiang et al.}

\institute{
MAIS, Institute of Automation, Chinese Academy of Sciences
\and
School of Artificial Intelligence, University of Chinese Academy of Sciences
\and
Tencent Hunyuan
\and 
HKUST
\and
Beijing Key Laboratory of Super Intelligent Security of Multi-Modal Information
\and
School of Information Science and Technology, ShanghaiTech University
\\
\email{jiangyanqin2021@ia.ac.cn; tengfeiwang12@gmail.com; jin.gao@nlpr.ia.ac.cn}
}

\maketitle

\authorfootnote{\dag}{Work done during an internship at Tencent Hunyuan.}
\authorfootnote{\ddag}{Tengfei Wang and Jin Gao are co-corresponding authors.}

{\centering
\url{https://omnix4d.github.io/}\par
}

\begin{abstract}
    Previous feed-forward 4D reconstruction methods either predict per-frame static point clouds, ignoring foreground motion, or estimate point cloud trajectories while being limited to small camera motions. This limits their ability to aggregate observations over time and reconstruct complete dynamic scenes under large viewpoint changes. To address this, we propose OmniX, a feed-forward 4D reconstruction framework that predicts \textbf{dense 3D point trajectories for every pixel} from videos with \textbf{large camera motion}. 1) OmniX separates dynamic motion modeling from static geometry prediction and represents motion with a small set of dynamic tokens. Leveraging the sparse and low-rank structure of 3D motion, these tokens generate trajectory fields for all pixels in all images while efficiently preserving global interactions. 2) To facilitate training, we build an automatic UE5-based 4D data engine and introduce a dataset of 80k scenes and 1.28M multi-view videos with full geometric annotations. OmniX achieves state-of-the-art results on dense 3D point trajectory prediction and 3D point tracking, with competitive performance on video depth estimation and camera pose estimation.
\end{abstract}

\section{Introduction}
Given image sequences of a dynamic scene, 4D reconstruction aims to reconstruct a dynamic 3D representation that is consistent across views and time. Serving as a foundation for dynamic scene understanding, it enables a wide range of downstream applications, including embodied AI simulation, autonomous driving perception, and AR/VR content creation. Thus, this area has attracted growing interest from the research community. 

Recent progress in 4D reconstruction has been largely driven by feed-forward approaches~\cite{monst3r,cut3r,page4d,vggt4d,st4rtrack,spatialtrackerv2,traceanything,vdpm}. Most existing methods predict per-frame 3D point clouds from videos~\cite{monst3r, cut3r, vggt4d, page4d, worldmirror, hyworld2, da3, pi3}, without explicitly modeling motion across time. To recover temporal correspondences, motion is often inferred by iteratively querying and tracking points across video frames~\cite{spatialtracker,cotracker,delta, deltav2, spatialtrackerv2, tapip3d}, which is computationally expensive and difficult to scale to dense predictions. A few works attempt to predict dense 3D point trajectories~\cite{dpm, st4rtrack, traceanything,vdpm}, but are typically limited to monocular videos with small camera motion. These methods struggle under large viewpoint changes because static geometry feature matching and dynamic foreground temporal correspondence learning are performed without disentanglement, thereby distracting from each other.

In this work, we present \textbf{Omni$\mathbb{X}$}, a 4D reconstruction framework that \textit{predicts dense 3D point trajectories from images captured at arbitrary viewpoints and times}. It is robust to large camera motion and temporally discontinuous inputs by explicitly disentangling dynamic foreground motion from static scene structure. This is achieved by our proposed \textbf{Sparse Spatiotemporal Attention} (SSA) mechanism, which shows that, given the sparse and low-rank structure of 3D motion, a small set of dynamic tokens can effectively parameterize trajectory fields for all pixels across views and time. To predict per-pixel trajectories while keeping dynamic token selection in SSA differentiable, we further introduce a \textbf{Deformable Trajectory Sampling Head} (DTSH). Together, these designs enable efficient global attention and produce 3D point trajectories for all pixels in all images across all timestamps in one pass.

For training, due to the lack of  4D scene datasets with large camera motion and 360° multi-view coverage, we develop an automatic data engine based on Unreal Engine 5 (UE5). It synthesizes dynamic scenes from static environments and dynamic objects, and renders them with a customized camera system. Using this engine, we construct a dataset with \textbf{80K scenes and 1.28M multi-view videos}, annotated with depth, camera poses, and dense 3D point trajectories.

OmniX achieves state-of-the-art performance on 4D point cloud reconstruction and 3D point tracking, while delivering competitive performance on video depth estimation and camera pose estimation. Besides, we achieve an optimal accuracy-efficiency trade-off. These results consistently demonstrate the effectiveness of the proposed reconstruction framework and data generation pipeline.
\section{Related Work}
\subsection{Feedforward 3D and 4D Reconstruction}
The pinoreer work of feedfoward 3D reconstruction is DUSt3R~\cite{dust3r}, which introduces dense point map prediction from unposed images. Subsequent works significantly improve scalability through all-to-all attention~\cite{fast3r, vggt, mapanything,worldmirror, pi3, da3,hyworld2}. Particularly, several recent works leverage dynamic-scene videos as training data and formulate 4D reconstruction as predicting per-frame 3D point clouds~\cite{monst3r,page4d,vggt4d,pi3,worldmirror,da3,hyworld2, vggt_long, streaming,any4d}. However, these methods primarily reconstruct per-frame 3D geometry without explicitly establishing point-level correspondences across time. In contrast, our method augments per-frame point-cloud prediction with temporally consistent dense 3D point trajectories in a feed-forward manner.

\subsection{3D Point Tracking}
Early works achieve 3D point tracking through test-time optimization~\cite{omnimotion} or by lifting 2D tracks to 3D~\cite{tapvid3d}. SpatialTracker~\cite{spatialtracker} introduces the first feed-forward 3D point tracker, encoding frames into triplane representations and iteratively updating point trajectories in 3D space. Subsequent methods follow this iterative tracking paradigm, improving scalability~\cite{spatialtrackerv2}, dense prediction~\cite{delta,deltav2}, and feature matching~\cite{tapip3d}.
Recent methods~\cite{dpm,st4rtrack,traceanything,vdpm} directly regress dense 3D point trajectories, substantially improving efficiency. The former two use pairwise prediction, while the latter two predict trajectories for all pixels across all frames. However, these methods mainly adapt the point prediction heads of 3D reconstruction models for trajectory prediction, without explicitly disentangling dynamic motion from static geometry. Another work D4RT~\cite{d4rt} proposes a flexible querying mechanism that removes both iterative trajectory updates and self-attention among queries, achieving impressive performance, but the model is not open-sourced.
Concurrent works include 4RC~\cite{4rc} and Track4World~\cite{track4world}. 4RC disentangles static and dynamic prediction but does not employ sparse attention. Track4World improves the efficiency of the classic iterative tracking architecture, achieveing high precision, but its inference speed remains limited by iterative updates.
In contrast, our method directly regresses dense 3D point trajectories while leveraging sparse attention and large-scale training data, enabling efficient and robust 3D point tracking under large camera motion.

\section{Method}
We aim at recovering consistent 4D geometry from any-view and any-time visual inputs, spanning the \textbf{single} video input, \textbf{multi-video} input and even the image-video \textbf{combinations}.

\subsection{Formulation}\label{sec:formulation}

We denote the input as a collection of videos $\mathcal{V}=\{\mathcal{V}^{(i)}\}_{i=1}^{N_v}$. The $i$-th video is represented by $\mathcal{V}^{(i)}=\{\mathbf{I}^{(i)}_j\}_{j=1}^{T^{(i)}}$, where $\mathbf{I}^{(i)}_{j} \in \mathbb{R}^{H \times W \times 3}$ denotes the $j$-th image in the $i$-th video and $T^{(i)}$ denotes the video length.

\paragraph{3D Geometry.} Each image $\mathbf{I}^{(i)}_{j}$ is associated with a depth map $\mathbf{D}^{(i)}_j \in \mathbb{R}^{H \times W}$,  a camera ray map $\mathbf{R}^{(i)}_j \in \mathbb{R}^{H \times W \times 3}$, the camera extrinsics $[\mathbf{Q}^{(i)}_j \mid \mathbf{t}^{(i)}_j]$ and intrinsics $\mathbf{K}^{(i)}_j$, where $H$ and $W$ are image height and width.
Alternatively, we represent the camera parameters as $\mathbf{v}_j = (\mathbf{t}_j, \mathbf{q}_j, \mathbf{f}_j) \in \mathbb{R}^9$, where $\mathbf{t}_j \in \mathbb{R}^3$ denotes camera translation, $\mathbf{q}_j \in \mathbb{R}^4$ is a rotation quaternion, and $\mathbf{f}_j \in \mathbb{R}^2$ represents the field-of-view parameters.
Following prior work, for each pixel $\mathbf{p}$, we define a camera ray $\mathbf{r} = (\mathbf{t}, \mathbf{d}) \in \mathbb{R}^6$. Here, $\mathbf{t}$ and $\mathbf{d}$ denote the ray origin and direction in world coordinates respectively. The 3D point $\mathbf{P}$ for pixel location $(u,v)$ is computed as 
\begin{equation}
    \mathbf{P} = \mathbf{t} + \mathbf{D}(u,v) \cdot \mathbf{d}.
\end{equation}

\paragraph{4D Geometry.}
We also predict the trajectory of each 3D point over the timestamps of the input observations.
Let $\mathbf{P}\equiv \mathbf{P}^{(i_0)}_{j_0}$ be a reference 3D point observed at time
$(i_0,j_0)$. We define its trajectory over the whole collection as
$\{\mathbf{P}^{(i)}_j\}$.
Inspired by Shape-of-Motion\cite{shape_of_motion}, we obtain this trajectory by a trajectory transformation $\boldsymbol{\tau}\;=\;\big\{[\mathbf{A}^{(i)}_j \mid \mathbf{b}^{(i)}_j]\big\}$, where $\mathbf{A}^{(i)}_j$ and $\mathbf{b}^{(i)}_j$ denote the rotation and translation matrices.
The transformed point at video $i$ and frame $j$ is
\begin{equation}\label{eq: pt3d_traj}
\mathbf{P}^{(i)}_j = \mathbf{A}^{(i)}_j\,\mathbf{P}^{(i_0)}_{j_0} + \mathbf{b}^{(i)}_j .
\end{equation}
Driven by the inherent sparsity of motion, we represent motion via \textbf{trajectory field} to ensure computational efficiency. The trajectory field, denoted as
$\mathbf{T}^{(i_0)}_{j_0}=\{\boldsymbol{\tau}^{(i_0)}_{j_0,k}\}_{k=1}^{K}$, serves as a set of \textbf{trajectory transformation bases}
for representing $\boldsymbol{\tau}$ in reference image $\mathbf{I}^{(i_0)}_{j_0}$.
That is, we construct the per-pixel trajectory transformation $\boldsymbol{\tau}$ by a weighted combination:
\begin{equation}
\boldsymbol{\tau} = \sum_{k=1}^{K} w_k\,\boldsymbol{\tau}^{(i_0)}_{j_0,k},
\end{equation}
where $\{w_k\}$ are learned combination weights predicted from $\mathbf{I}^{(i_0)}_{j_0}$.

\begin{figure}[htbp]
    \centering
    \includegraphics[width=0.98\textwidth]{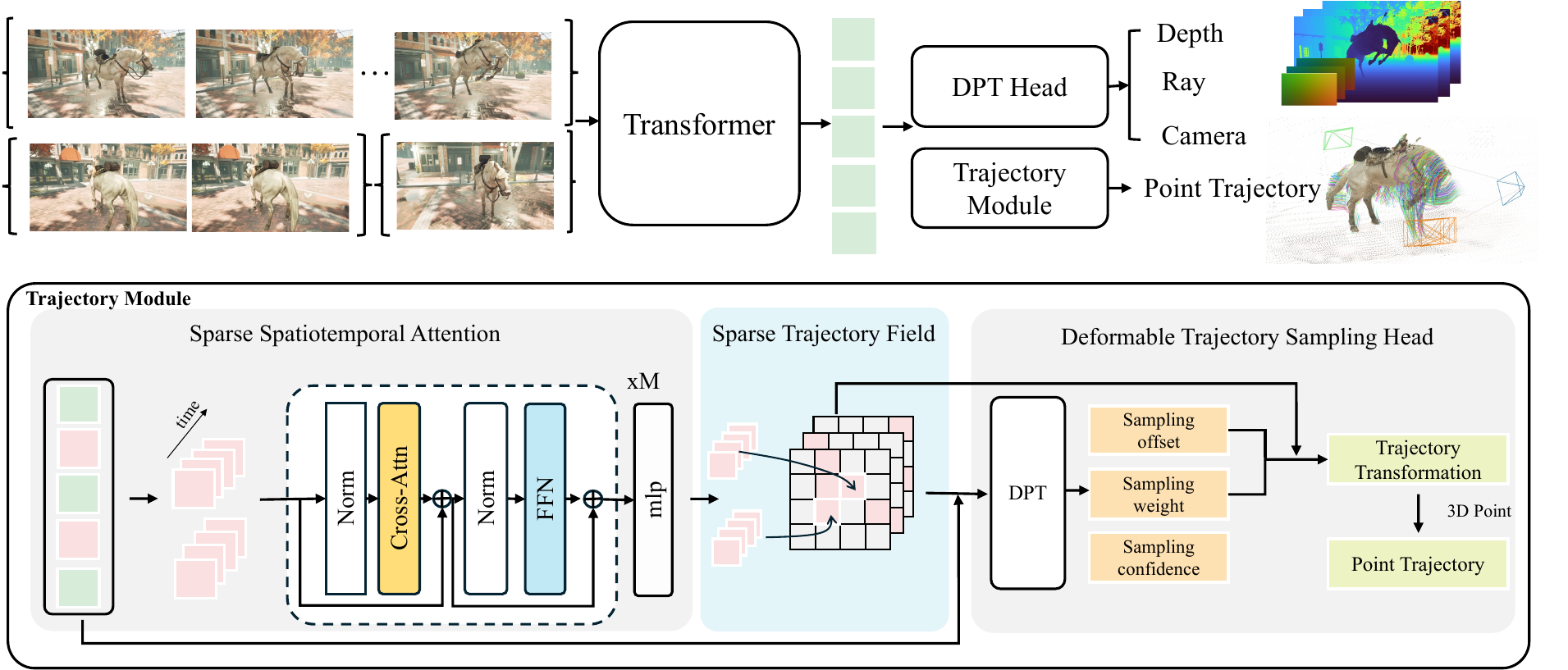} 
    \caption{Framework of OmniX. Built upon DepthAnything3~\cite{da3}, OmniX predicts depth, ray maps, and camera parameters for the input images, from which 3D points are recovered. A trajectory module then predicts transformations for these 3D points. For efficiency, we propose a Sparse Spatiotemporal Attention (SSA) mechanism, which selects dynamic tokens, expands them across timestamps as queries, and lets them cross-attend to all image tokens to predict trajectory transformation bases. These bases are scattered back to the token grid to form a sparse trajectory field. A deformable trajectory sampling head upsamples this field to pixel-level transformation maps, which are applied to the 3D points to obtain dense trajectories.}
    \label{fig:omnix_pipeline}
\end{figure}
\subsection{Architecture}

\paragraph{Overview.} 
As shown in~\cref{fig:omnix_pipeline}, building upon the feed-forward 3D reconstruction models~\cite{vggt,worldmirror,da3}, OmniX consists of three main components: a transformer backbone, a DPT head for 3D geometry prediction, and a newly introduced trajectory module for 4D geometry prediction. Given any-view and any-time visual inputs, the transformer backbone encodes all images into tokens, injects sinusoidal encodings to embed timestamps for all images, and performs cross-view self-attention to aggregate multi-view information. The DPT head then predicts depth and ray maps for each image, from which 3D points are recovered.
To lift 3D geometry to 4D, we introduce a trajectory module composed of a Sparse Spatiotemporal Attention (SSA) mechanism and a Deformable Trajectory Sampling Head (DTSH). SSA selects a set of dynamic tokens and expands them along the temporal dimension as trajectory queries. These expanded tokens are processed by stacked spatiotemporal cross-attention blocks, where they attend to all image tokens to establish spatiotemporal correspondences and predict trajectory transformation bases. The predicted bases are scattered back to their spatial token locations, forming a sparse trajectory field. Given this field, DTSH samples neighboring bases for each pixel and combines them with the learned weights to produce pixel-level trajectory transformations, which lift the recovered 3D points to dense 4D trajectories. We next detail SSA, the sparse trajectory field construction, DTSH, and the differentiable dynamic token selection mechanism.

\paragraph{Sparse Spatiotemporal Attention.}
Given the image tokens $\mathbf{F}^{(i)}_j \in \mathbb{R}^{L \times d}$ from the $j$-th reference image of the $i$-th video, we employ an MLP $\varphi$ to predict the dynamic probability $\boldsymbol{\Theta}^{(i)}_j$ and select the top-$\rho\%$ tokens as dynamic tokens, denoted as $\hat{\mathbf{F}}^{(i)}_j \in \mathbb{R}^{\hat{L} \times d}$. $L$ and $\hat{L}$ are the number of image and dynamic tokens, respectively. Dynamic tokens are processed to predict trajectory transformation bases.
We first expand them along temporal dimension by temporal embedding. 
The temporal embedding is constructed by adding the sinusoidal encodings of the video index and the local frame index, denoted by $\mathbf{e} \in \mathbb{R}^{\mathcal{T} \times d}$, where
$\mathcal{T} = \sum_{i=1}^{N_v} T^{(i)}$ is the total number of input images. 
To capture temporal dynamics more effectively, we modulate the temporal embedding with per-token coefficients predicted from the dynamic tokens. 
The temporally expanded dynamic tokens
$\tilde{\mathbf{F}}^{(i)}_j \in \mathbb{R}^{\mathcal{T} \times \hat{L} \times d}$ are computed as
\begin{equation}\label{eq:temporal embed}
     \tilde{\mathbf{F}}^{(i)}_j
    =
    \hat{\mathbf{F}}^{(i)}_j
    + \gamma(\hat{\mathbf{F}}^{(i)}_j) \cdot \beta(\mathbf{e}),
\end{equation}
where $\gamma$ and $\beta$ are MLPs with outputs
$\gamma(\hat{\mathbf{F}}^{(i)}_j) \in \mathbb{R}^{\hat{L} \times 1}$ and
$\beta(\mathbf{e}) \in \mathbb{R}^{\mathcal{T} \times d}$, respectively. The \texttt{broadcasting} along the temporal and token dimensions is omitted for simplicity.

We collect $\tilde{\mathbf{F}}^{(i)}_j$ and $\mathbf{F}^{(i)}_j$ from all input images and stack them along the first dimension, obtaining
$\tilde{\mathbf{F}} \in \mathbb{R}^{\mathcal{T} \times \mathcal{T} \times \hat{L} \times d}$ and
$\mathbf{F} \in \mathbb{R}^{\mathcal{T} \times L \times d}$.
In the spatiotemporal cross-attention module, $\tilde{\mathbf{F}}$ serves as queries to extract geometric and motion cues from $\mathbf{F}$. Self-attention among query tokens is omitted, since experiments in~\cref{sec:ablation} indicates that it provides negligible benefit in this module. For the $m$-th attention block, let the output be $\mathbf{Z}^m_c$ and the intermediate output be $\mathbf{Z'}^m_c$. The processing flow is formulated as follows:
\begin{equation}
    \mathbf{Z'}^m_c = \mathrm{SSA}^m_c \Big( \mathrm{LN}(\mathbf{Z}^{m-1}_c), \mathbf{F} \Big) + \mathbf{Z}^{m-1}_c,
\end{equation}
where $\mathbf{Z}^0_c = \tilde{\mathbf{F}}$, and $\mathrm{LN}$ denotes layer normalization. The $\mathrm{SSA}^m_c$ operator aggregates spatiotemporal information through a multi-head cross-attention mechanism. For clarity, the core scaled dot-product attention operation is defined as follows:
\begin{equation}
    \mathrm{SSA}^m_c(\mathbf{Z}^{\mathrm{r}}, \mathbf{F}^{\mathrm{r}}) = \mathrm{Softmax} \left( \frac{(\mathbf{Z}^{\mathrm{r}} \mathbf{W}^m_Q)(\mathbf{F}^{\mathrm{r}} \mathbf{W}^m_K)^	T}{\sqrt{d}} \right) (\mathbf{F}^{\mathrm{r}} \mathbf{W}^m_V).
\end{equation}
Here, $\mathbf{Z}^{\mathrm{r}} \in \mathbb{R}^{\tilde{\mathcal{l}} \times d}$ and $\mathbf{F}^{\mathrm{r}} \in \mathbb{R}^{\mathcal{l} \times d}$ are the reshaped output of the queries and image tokens. $\mathbf{W}^m_Q, \mathbf{W}^m_K, \mathbf{W}^m_V \in \mathbb{R}^{d \times d}$ represent the learnable projection matrices for generating queries, keys, and values in the $m$-th block, respectively.
The cross-attention output is further processed by a feed-forward network, denoted as $\mathrm{FFN}$, along with residual connections and layer normalization, as follows:
\begin{equation}
    \mathbf{Z}^m_c = \mathbf{Z'}^{m}_c + \mathrm{FFN} \Big( \mathrm{LN}(\mathbf{Z}'^m_c) \Big).
\end{equation}
Through this cascaded cross-attention architecture, the model can adaptively retrieve and integrate key geometric constraints and motion cues from the global image features $\mathbf{F}$, guided by the spatiotemporal priors in $\tilde{\mathbf{F}}$, thereby obtaining highly consistent trajectory representations. 

\paragraph{Sparse Trajectory Field.}
The trajectory field $\mathbf{T} \in \mathbb{R}^{\mathcal{T} \times \mathcal{T} \times \hat{L} \times c}$ is predicted as follows:
\begin{equation}
   \mathbf{T}^{m} = \phi^m(\mathbf{Z}^m_c),
\end{equation}
where $\phi^m$ denotes a MLP layer. The channel dimension $c=9$, comprises two components: the first six channels represent the 6D rotation parameters, and the remaining three channels correspond to the 3D translation vector. 
To enable efficient sampling in subsequent stages, we \texttt{scatter} these trajectory transformations back into 2D spatial grids according to the original positions of the dynamic tokens, obtaining a scattered sparse trajectory field.
Finally, we reshape the field by merging the temporal dimension with the channel dimension to facilitate grid-based sampling operations, \ie, the sparse trajectory field $\tilde{\mathbf{T}}^m \in \mathbb{R}^{\mathcal{T} \times \mathcal{H} \times \mathcal{W} \times (\mathcal{T} \cdot c)}$, where $\mathcal{H} \times \mathcal{W} = L$.

\paragraph{Deformable Trajectory Sampling Head.}
To obtain per-pixel trajectory transformations from sparse trajectory fields, we design a deformable sampling head based on the DPT architecture~\cite{dpt}. While the original DPT head operates on multi-level image tokens, we augment these tokens by embedding both the trajectory fields and per-token dynamic scores. Specifically, we aggregate these sparse fields via two pathways: a) a statistical branch that computes the temporal mean and variance of the sparse trajectory field, followed by a MLP projection; and b) a salient branch that projects each field individually before applying temporal max pooling. These embeddings, combined with MLP-projected dynamic scores, are concatenated with the image tokens to generate the sampling parameters, \ie, sampling weights and sampling offsets.

To maintain the \textit{differentiability of dynamic token selection}, we adopt a multi-scale deformable sampling mechanism to jointly sample trajectory transformations and dynamic scores, generating the final trajectories via weighted fusion. Specifically, the head predicts sampling offsets $\Delta \mathbf{p}^m_q$ and weights $w^m_q$ for pixel $\mathbf{p}$, where q indexes the sampling point. Let $\mathcal{B}$ denotes bilinear sampling, we compute the per-pixel trajectory transformation $\boldsymbol{\tau}$ and dynamic score $\boldsymbol{\theta}$ as:
\begin{equation}
    \boldsymbol{\tau} = \sum^m\sum^q w^m_q \mathcal{B}(\mathbf{\tilde{\mathbf{T}}^m, \mathbf{p} +\Delta \mathbf{p}^m_q}),
\end{equation}
\begin{equation}
    \boldsymbol{\theta} = \sum^m\sum^q w^m_q \mathcal{B}(\mathbf{\boldsymbol{\Theta}, \mathbf{p} +\Delta \mathbf{p}^m_q}).
\end{equation}
Here, $\boldsymbol{\Theta} \in \mathbb{R}^{\mathcal{T} \times \mathcal{H} \times \mathcal{W}}$ is the stacked and rearranged result of dynamic probability $\boldsymbol{\Theta}^{(i)}_{j}$ in SSA, shared by all levels. We compute the point trajectory $\mathbf{P}_{traj} \equiv \{\mathbf{P}^{(i)}_j\}$ following Eq.~\eqref{eq: pt3d_traj}. To allow the trajectory loss to guide the selection of dynamic tokens, we incorporate the dynamic score $\boldsymbol{\theta}$ into the final trajectory prediction:
\begin{equation}
    \hat{\mathbf{P}}_{traj} = \mathbf{P} + \boldsymbol{\theta} \cdot ( \mathbf{P}_{traj} - \mathbf{P}).
\end{equation}
The \texttt{broadcasting} of $\mathbf{P}$ is omitted for simplicity. This trajectory formulation also encourages background points to remain static during prediction.

\subsection{Training Objectives}
Following the formulation in~\cref{sec:formulation}, our model maps the video collection $\mathcal{V}$ to a depth map $\mathbf{D}$, a ray map $\mathbf{R}$, an optional camera pose $\mathbf{c}$, and a trajectory map $\mathbf{T}$. Additionally, we predict a token-level dynamic score map $\boldsymbol{\Theta}$ and a pixel-level dynamic score map $\boldsymbol{\theta}$. Before loss computation, all ground-truth signals are normalized by the average distance of ground-truth point cloud trajectories, following prior works, to ensure magnitude consistency across different modalities and stabilize the training process. 

We supervise the depth, ray map, and point trajectory using confidence-aware losses. While the dynamic score maps do not theoretically require explicit supervision, we incorporate them with a small weighting factor for regularization when ground-truth dynamic masks are available. Let $\mathbf{D}^*$, $\mathbf{R}^*$, $\mathbf{c}^*$, and $\mathbf{T}^*$ denote the ground-truth counterparts of the predicted variables. The total objective function is defined as:
\begin{equation}
    \mathcal{L} = \mathcal{L}_D + \mathcal{L}_R + \mathcal{L}_{T} + \beta \mathcal{L}_C + \alpha \mathcal{L}_{grad} + \kappa (\mathcal{L}_{\Theta} + \mathcal{L}_{\theta}),
\end{equation}
where $\beta$, $\alpha$, and $\kappa$ are hyper-parameters balancing different tasks. For the geometric predictions $\mathbf{X} \in \{\mathbf{D}, \mathbf{R}, \mathbf{T}\}$, we incorporate a confidence-aware supervision mechanism. This allows the model to adaptively re-weight the regression loss based on the estimated uncertainty, thereby alleviating the impact of outliers or ambiguous regions (\eg, occlusions and sky region). Let $\mathbf{C}_{\mathbf{X}}$ denote the predicted confidence map corresponding to variable $\mathbf{X}$. The regression loss is formulated as:
\begin{equation}
    \mathcal{L}_{reg}(\mathbf{X}, \mathbf{X}^*) = \frac{1}{|\Omega|} \sum_{p \in \Omega} \left( (1 + \lambda_1 \mathbf{C}_{\mathbf{X}, p}) \| \mathbf{X}_p - \mathbf{X}^*_p \|_1 - \lambda_2 \log \mathbf{C}_{\mathbf{X}, p} \right),
\end{equation}
where $\Omega$ denotes valid pixels and $\lambda_1$ and $\lambda_2$ are hyperparameters balancing the weighted reconstruction error and the confidence regularization. This loss is applied as  follows:
\begin{equation}
\mathcal{L}_{D} = \mathcal{L}_{reg}(\mathbf{D}, \mathbf{D}^*), \mathcal{L}_{R} = \mathcal{L}_{reg}(\mathbf{R}, \mathbf{R}^*), \mathcal{L}_{T} = \mathcal{L}_{reg}(\mathbf{T}, \mathbf{T}^*).
\end{equation}
To preserve sharp edges while ensuring spatial consistency in planar regions, we introduce the gradient loss $\mathcal{L}_{grad}$ to penalize discrepancies in depth gradients:
\begin{equation}
    \mathcal{L}_{grad}(\mathbf{D}, \mathbf{D}^*) = \| \nabla_x \mathbf{D} - \nabla_x \mathbf{D}^* \|_1 + \| \nabla_y \mathbf{D} - \nabla_y \mathbf{D}^* \|_1,
\end{equation}
where $\nabla_x$ and $\nabla_y$ are the horizontal and vertical finite difference operators, respectively.
For camera pose optimization, to ensure robustness against outliers, we apply the Huber loss $|\cdot|_{\epsilon}$:
\begin{equation}
    \mathcal{L}_C = |f - f^*|_{\epsilon} + |\mathbf{q} - \mathbf{q}^*|_{\epsilon} + |\mathbf{t} - \mathbf{t}^*|_{\epsilon},
\end{equation}
where $|\cdot|_{\epsilon}$ is the Huber loss with threshold $\epsilon$. 
Regarding the dynamic score maps, we employ the Binary Cross-Entropy (BCE) loss for regularization at both token and pixel granularities, whereas all other loss terms are formulated based on the $\ell_1$ norm. Let $\mathbf{M} \in \{\boldsymbol{\Theta}, \boldsymbol{\theta}\}$ represent a dynamic score map at a specific scale and $\Omega_{\mathbf{M}}$ be its corresponding spatial domain. The BCE loss is defined as:
\begin{equation}
    \mathcal{L}_{\text{BCE}}(\mathbf{M}, \mathbf{M}^*) = - \frac{1}{|\Omega_{\mathbf{M}}|} \sum_{p \in \Omega_{\mathbf{M}}} \left[ \mathbf{M}^*_p \log \mathbf{M}_p + (1 - \mathbf{M}^*_p) \log (1 - \mathbf{M}_p) \right].
\end{equation}
Thus, 
\begin{equation}
    \mathcal{L}_{\Theta} = \mathcal{L}_{\text{BCE}}(\boldsymbol{\Theta}, \boldsymbol{\Theta}^*),~~ \mathcal{L}_{\theta} = \mathcal{L}_{\text{BCE}}(\boldsymbol{\theta}, \boldsymbol{\theta}^*).
\end{equation}
The joint optimization of loss terms of different tasks facilitates the understanding of the dynamic 3D scene.

\section{Experiment}
\subsection{Setup}

\noindent\textbf{Data Preparation.} 
To address the scarcity of large-scale 4D scene datasets featuring full geometric annotations and multi-view configurations, we develop an automated data generation engine based on Unreal Engine 5~\cite{unrealengine5}. This pipeline constructs dynamic scenes by compositing static environments with dynamic objects. 
A multi-camera system with diverse camera motion is designed to capture the dynamic scenes. We construct 160k dynamic scenes and render each with 16 cameras. After data cleaning, the dataset contains 80K scenes and approximately 1.28M multi-veiw videos, with 77K scenes used for training and evaluation in this work.

\noindent\textbf{Implementation Details.}
In addition to our synthetic data, we incorporate eight public datasets to enhance diversity: DynamicReplica~\cite{dynamicstereo}, PointOdessey\cite{pointodyssey}, Spring~\cite{spring}, OmniGame~\cite{omniworld}, HOI4D~\cite{hoi4d}, Waymo~\cite{waymo}, DL3DV~\cite{dl3dv}, and Stereo4D~\cite{stereo4d}. 
The model is configured with 8 SSA blocks. The weight for the dynamic mask loss is set to 0.01. During training, we use image resolution of $280\times504$ and sequence length of 16 frames; however, we observe that the model can generalize to longer sequences (\eg, 32 frames or more) during inference. 
We employ the AdamW optimizer~\cite{adamw} with a learning rate of 2e-4. The model is trained for approximately 64k steps on a cluster of 64 GPUs, with a batch size of 1 per GPU. The total training time is approximately 12 days.

\noindent\textbf{Evaluation.}
We establish a comprehensive benchmark on the validation set of the proposed dataset to evaluate per-pixel and per-frame 3D point trajectory prediction, referred to as dense 3D point trajectory prediction. The evaluation data consist of 600 videos from 40 scenes.
This benchmark supports diverse input types, including monocular videos with complex camera motions, temporally disjoint video pairs, and hybrid image-video sets, and provides ground truth of dense 3D point trajectories.
Following St4rTrack~\cite{st4rtrack}, we report the End-Point Error (EPE) and Average Point Distance (APD3D), both evaluated in the world coordinate and averaged by trajectories from all frames. We compute the two metrics for foreground points (FG) and all points (ALL) independently.

We also evaluate our model on sparse 3D point tracking task, using the TAPVid-3D dataset\cite{tapvid3d}.
We further compare our method with state-of-the-art approaches on video depth estimation and camera pose estimation tasks, utilizing the KITTI~\cite{kitti}, Sintel~\cite{sintel}, and TUM-dynamic~\cite{tum_dynamic} datasets.
At last, we evaluate the efficiency of the proposed framework.
For all evaluation tasks, videos are clipped into 16-frame sequences to keep evaluation within the sequence length used during the training of our model and also by other dense 3D point trajectory prediction methods~\cite{traceanything,vdpm}.

\subsection{Comparison with State-of-the-arts}
\begin{table}[htbp]
\centering
\caption{Dense 3D point trajectory prediction evaluation. The evaluated models are trained on different data sources. TraceAnything is trained on data generated by its custom engine. VDPM is trained on 4 datasets, including the unreleased Kubric-G~\cite{dpm}. Our model is trained on 9 datasets, including our custom UE dataset.}
\label{tab:omnix_4d_traj}
\resizebox{\textwidth}{!}{
\begin{tabular}{l c c c c c c c c c c c c c}
\toprule
\multirow{3}{*}{\textbf{Method}} & \multirow{3}{*}{\textbf{Eval Type}} & \multicolumn{4}{c}{\textbf{Monocular Video}} & \multicolumn{4}{c}{\textbf{Disjoint Video Pairs}} & \multicolumn{4}{c}{\textbf{Hybrid Image-video sets}} \\
\cmidrule(lr){3-6} \cmidrule(lr){7-10} \cmidrule(lr){11-14}
& & \multicolumn{2}{c}{APD3D$\!$ $\uparrow$} & \multicolumn{2}{c}{EPE$\!$ $\downarrow$} & \multicolumn{2}{c}{APD3D$\!$ $\uparrow$} & \multicolumn{2}{c}{EPE$\!$ $\downarrow$} & \multicolumn{2}{c}{APD3D$\!$ $\uparrow$} & \multicolumn{2}{c}{EPE$\!$ $\downarrow$} \\
\cmidrule(lr){3-4} \cmidrule(lr){5-6} \cmidrule(lr){7-8} \cmidrule(lr){9-10} \cmidrule(lr){11-12} \cmidrule(lr){13-14}
& & FG & ALL & FG & ALL & FG & ALL & FG & ALL & FG & ALL & FG & ALL \\
\midrule
TraceAnything~\cite{traceanything} & \multirow{3}{*}{first frame} & 0.062 & 0.043 & 0.354 & 0.316 & 0.059 & 0.037 & 0.380 & 0.343 & 0.030 & 0.027 & 0.521 & 0.447 \\
VDPM~\cite{vdpm}          &                              & 0.187 & 0.120 & 0.338 & 0.287 & 0.164 & 0.114 & 0.341 & 0.306 & 0.146 & 0.117 & 0.458 & 0.405 \\
\textbf{Ours} &                              & \textbf{0.284} & \textbf{0.391} & \textbf{0.133} & \textbf{0.116} & \textbf{0.322} & \textbf{0.444} & \textbf{0.101} & \textbf{0.094} & \textbf{0.332} & \textbf{0.456} & \textbf{0.146} & \textbf{0.133} \\
\midrule
VDPM~\cite{vdpm}          & \multirow{2}{*}{all frames}  & 0.199 & 0.118 & 0.332 & 0.282 & 0.143 & 0.093 & 0.345 & 0.305 & 0.135 & 0.104 & 0.341 & 0.306 \\
\textbf{Ours} &                              & \textbf{0.300} & \textbf{0.381} & \textbf{0.133} & \textbf{0.116} & \textbf{0.305} & \textbf{0.406} & \textbf{0.113} & \textbf{0.105} & \textbf{0.316} & \textbf{0.400} & \textbf{0.132} & \textbf{0.121} \\
\bottomrule
\end{tabular}
}
\end{table}
\begin{table*}[!htb] 
\centering
\caption{3D point tracking evaluation on the TAPVid-3D benchmark~\cite{tapvid3d}. Input videos are segmented into 16-frame clips. The evaluated models are trained on different data sources. SpatialTrackerV2 is trained on 17 datasets, including both synthetic and real-world datasets. St4RTrack is trained on 3 synthetic datasets. TraceAnything is trained on data generated by its custom engine. VDPM is trained on 4 datasets, including the unreleased Kubric-G~\cite{dpm}. Our model is trained on 9 datasets, including our UE dataset. Note that VDPM is trained on DriveTrack (Waymo) data, which may overlap with the test data, while we exclude the specific test scenes during training.}
\label{tab:omnix_3d_tracking}
\resizebox{0.8\textwidth}{!}{ 
\begin{tabular}{l c c c c c c}
\toprule
\multirow{2}{*}{\textbf{Method}} & \multicolumn{2}{c}{\textbf{ADT}\cite{adt}} & \multicolumn{2}{c}{\textbf{DriveTrack}\cite{waymo}} & \multicolumn{2}{c}{\textbf{PStudio}\cite{pstudio}} \\
\cmidrule(lr){2-3} \cmidrule(lr){4-5} \cmidrule(lr){6-7}
 & \textbf{APD3D}$\!$ $\uparrow$ & \textbf{EPE}$\!$ $\downarrow$ & \textbf{APD3D}$\!$ $\uparrow$ & \textbf{EPE}$\!$ $\downarrow$ & \textbf{APD3D}$\!$ $\uparrow$ & \textbf{EPE}$\!$ $\downarrow$ \\
\midrule
SpatialTrackerV2\cite{spatialtrackerv2} & 0.236 & 0.193 & 0.197 & 1.651 & 0.107 & \underline{0.149} \\
St4RTrack\cite{st4rtrack} & 0.202 & 0.259 & 0.124 & 1.906 & 0.099 & 0.191 \\
TraceAnything\cite{traceanything} & 0.069 & 0.297 & 0.113 & 2.638 &  0.097 & 0.189 \\
VDPM\cite{vdpm} & \underline{0.266} & \underline{0.162} & \underline{0.228} & \underline{1.576} & \underline{0.118} & \textbf{0.141} \\
\textbf{Ours} & \textbf{0.367} & \textbf{0.134} & \textbf{0.256} & \textbf{1.490} & \textbf{0.141} & 0.156 \\
\bottomrule
\end{tabular}
}
\end{table*}
\subsubsection{Dense 3D Point Trajectory Prediction.}
\cref{tab:omnix_4d_traj} reports the evaluation results of dense 3D point trajectory prediction on the validation set of our proposed dataset. Since TraceAnything~\cite{traceanything} does not predict camera parameters, we can only compute its metrics on the first frame rather than over the full sequence. In contrast, VDPM~\cite{vdpm} predicts both camera parameters and dense 3D point trajectories, allowing comparison under the full evaluation setting.
Our method outperforms both compared methods by a large margin under challenging input settings, especially for \textit{Disjoint Video Pairs} and \textit{Hybrid Image-video Sets}. This is mainly because the compared methods are not trained on videos with large camera motion, whereas our model explicitly includes such data during training. The significant performance gains on ADT~\cite{adt}, a dataset featuring large camera motion, in~\cref{tab:omnix_3d_tracking} further validate this point. We also observe that the foreground-point results of existing methods~\cite{traceanything,vdpm} are better than their all-point results. This is because their inaccurate camera estimation often causes many background points to be projected out of view during metric computation, even after scale alignment, whereas foreground points are typically more spatially concentrated and are therefore less likely to be projected out of view.

\begin{figure}[htbp]
    \centering
    \includegraphics[width=0.96\linewidth]{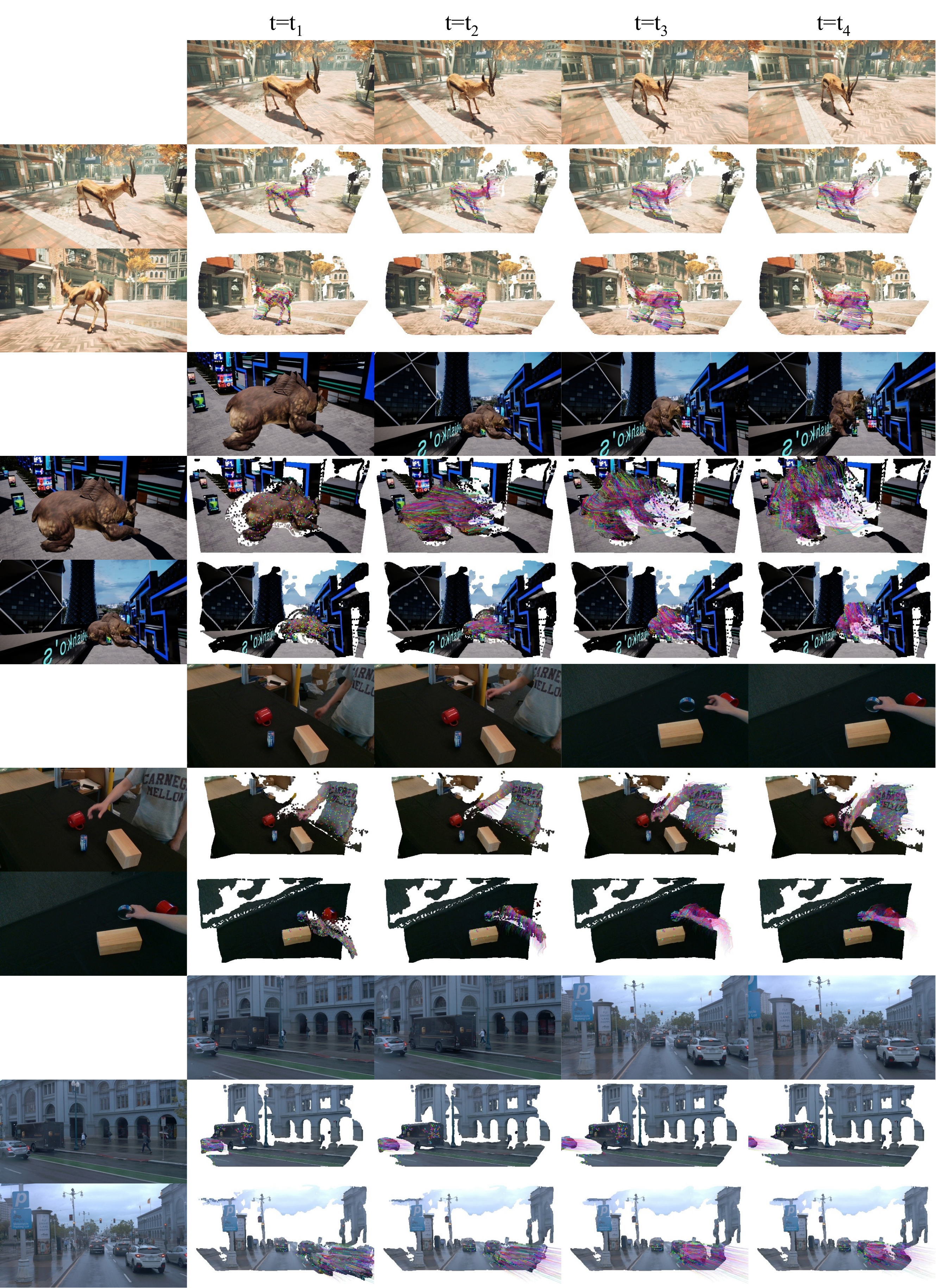}
    \caption{Visualization of dense 3D point trajectory prediction under challenging camera motion. The first two scenes are from our custom dataset, and the last two row features a scene from the DexYCB~\cite{dexycb} and Waymo~\cite{waymo} datasets. Each scene is shown as a $3\times5$ grid, where the top row presents input images at different timesteps. The first column of the second and third rows shows input images from different viewpoints, and columns 2--5 in the same row visualize the predicted trajectories of points predicted from the corresponding first-column image, at the timesteps indicated by the top row.}
    \label{fig:omnix_4d}
\end{figure}

Due to space limitations, we omit qualitative comparisons here and instead provide visualizations of our method in~\cref{fig:omnix_4d}, covering both synthetic and real-world dynamic scenes. As shown in~\cref{fig:omnix_4d}, our method can reconstruct dynamic scenes from inputs with camera motions of up to 180\textdegree{} (1st scene), while accurately predicting 3D point trajectories under both non-rigid and rigid motion.
\vspace{-12pt}

\subsubsection{Sparse 3D Point Tracking.}
\begin{figure}[!htb]
    \centering 
    \includegraphics[width=0.96\textwidth]{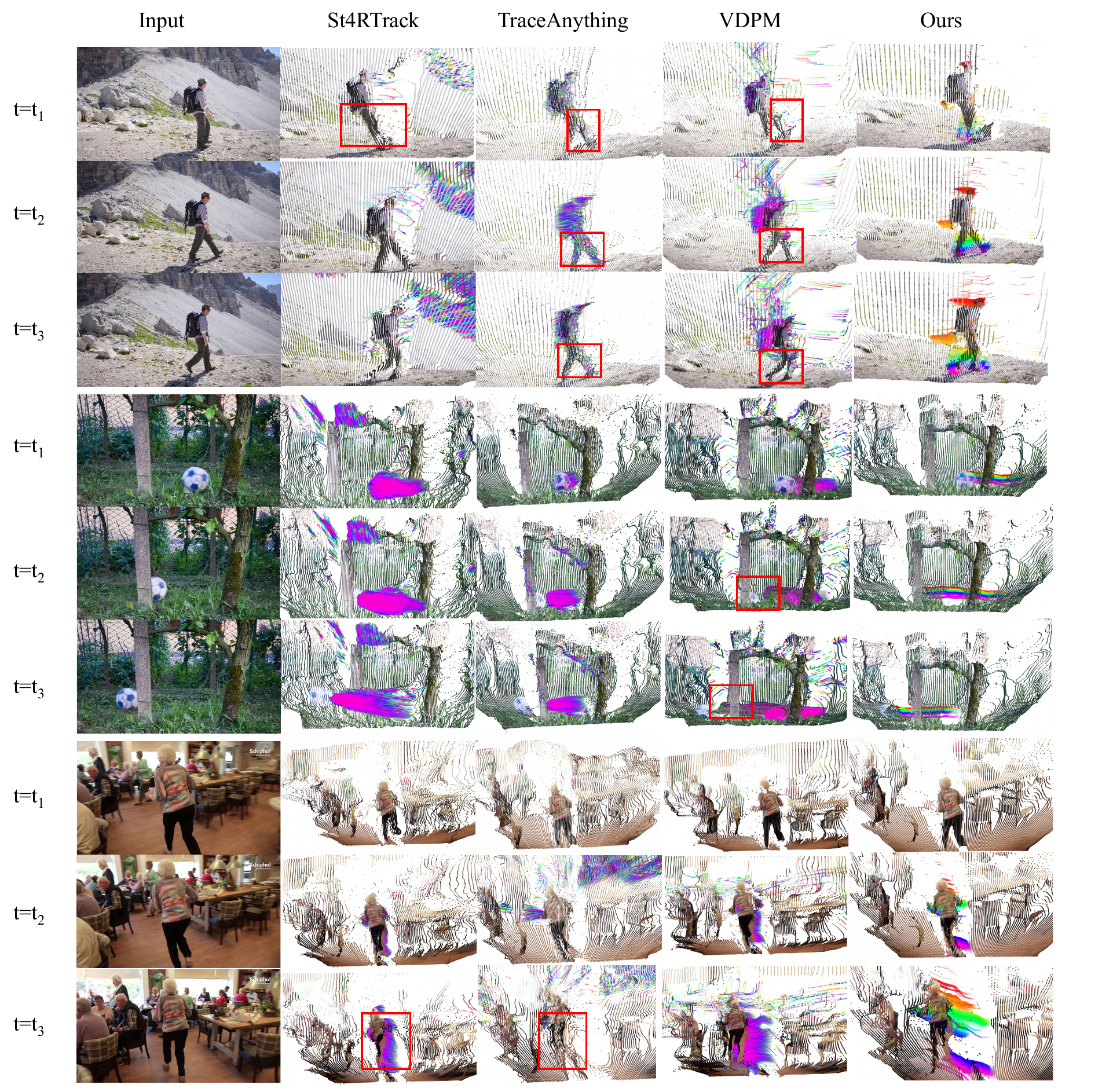} 
    \caption{Qualitative comparison of 3D point tracking on the DAVIS dataset~\cite{davis}. } 
    \label{fig:3d_tracking_comparison} 
    \vspace{-10pt}
\end{figure}

In~\cref{tab:omnix_3d_tracking}, we report evaluation results on the TAPVid-3D benchmark~\cite{tapvid3d}. Following standard protocols, videos are segmented into continuous 16-frame clips, and we evaluate all visible points starting from the first frame of each clip. Our method achieves state-of-the-art performance across all three datasets. Qualitative comparisons are conducted on the commonly used DAVIS dataset~\cite{davis} as shown in~\cref{fig:3d_tracking_comparison}. For visualization, we select points with the largest motion amplitudes while retaining background points, as background regions are also essential for dense reconstruction. We observe that: 1) In complex walking scenes, such as the first scene, comparison methods often fail to capture leg-crossing motion, resulting in tilting and warping artifacts. These methods tend to overfit to the object's static shape and fail to recover its actual motion. In contrast, our method correctly captures the forward-swinging and crossing motion patterns of the legs. 2) Comparison methods often fail to distinguish foreground objects from the static background and thus assign spurious displacements to background points. In contrast, our per-point dynamic score prediction better separates dynamic foreground points from the static background, reducing such false background displacements.
\vspace{-6pt}

\begin{table}[t]
\centering
\caption{Video depth estimation on the KITTI dataset\cite{kitti} (16-frame clips).}
\label{tab:omnix_video_depth}
\resizebox{\linewidth}{!}{
\begin{tabular}{l c c c c c c}
\toprule
\textbf{Metric} & VGGT4D$\!\!$~\cite{vggt4d} & PAGE4D$\!\!$~\cite{page4d} & $\pi^3$$\!\!$~\cite{pi3} & SpatialTrackerV2$\!\!$~\cite{spatialtrackerv2} & DepthAnythingV3$\!\!$~\cite{da3} & \textbf{Ours} \\
\midrule
Abs Rel$\!$ $\downarrow$ & 0.041 & \underline{0.031} & 0.032 & 0.049 & 0.039 & 	\textbf{0.024} \\
$\delta < 1.25\! \uparrow$ & 0.972 & 0.982 & 	\textbf{0.990} & 0.981 & 0.987 & \textbf{0.990} \\
\bottomrule
\end{tabular}
} 
\end{table}

\begin{table}[t]
\centering
\caption{Camera pose estimation (16-frame clips).}
\label{tab:omnix_camera_pose}
\resizebox{0.9\textwidth}{!}{
\begin{tabular}{l ccc ccc}
\toprule
\multirow{2}{*}{\textbf{Method}} & \multicolumn{3}{c}{\textbf{Sintel}\cite{sintel}} & \multicolumn{3}{c}{\textbf{TUM-dynamics}\cite{tum_dynamic}} \\
\cmidrule(lr){2-4} \cmidrule(lr){5-7}
& ATE $\downarrow$ & $\text{RPE}_{trans}$ $\downarrow$ & $\text{RPE}_{rot}$ $\downarrow$
& ATE $\downarrow$ & $\text{RPE}_{trans}$ $\downarrow$ & $\text{RPE}_{rot}$ $\downarrow$ \\
\midrule
VGGT4D\cite{vggt4d} & 0.321 & 0.322 & 1.165 & 0.011 & \underline{0.014} & \underline{0.461} \\
SpatialTrackerV2\cite{spatialtrackerv2} & 0.186 & 0.236 & 1.645 & 0.037 & 0.042 & 1.067 \\
PAGE4D\cite{page4d}  & 0.350 & 0.262 & 1.182 & 0.022 & 0.024 & 0.716 \\
VDPM\cite{vdpm} & 0.231 & 0.234 & 1.048 & 0.157 & 0.018 & 0.517 \\
$\pi^3$\cite{pi3} & \underline{0.157} & \underline{0.174} & \textbf{0.627} & 0.016 & 0.016 & 0.511 \\
DepthAnythingV3\cite{da3} & 0.252 & 0.299 & \underline{0.710} & \textbf{0.010} &\textbf{0.013} & \textbf{0.458} \\
\textbf{Ours} & \textbf{0.108} & \textbf{0.152} & 0.722 & \textbf{0.010} & \underline{0.014} & 0.472 \\
\bottomrule
\end{tabular}
}
\vspace{-8pt}
\end{table}

\subsubsection{Other Tasks.}
In addition to trajectory prediction, we evaluate our model on other tasks such as video depth estimation and camera pose estimation. Videos are segmented to 16-frame clips. As shown in~\cref{tab:omnix_video_depth} and~\cref{tab:omnix_camera_pose}, our method achieves comparable performance to current state-of-the-art methods. 

\subsubsection{Efficiency.}
\cref{tab:eff} compares the inference efficiency of different methods when evaluated on 16-frame clips. TraceAnything~\cite{traceanything} achieves the highest efficiency, attributed largely to its simple architecture, which only adds a tracking head to predict trajectory curves for pixels. However, its accuracy remains much lower than that of other methods, as shown in \cref{tab:omnix_4d_traj} and \cref{tab:omnix_3d_tracking}. Our method achieves the second-best efficiency, benefiting from the sparse spatiotemporal attention and a low-dimensional trajectory module (512-dimensional). Our method uses more memory because it natively predicts all trajectories in a single forward pass, rather than relying on an inference-time shortcut. In contrast, VDPM~\cite{vdpm} and St4RTrack~\cite{st4rtrack} are trained with sampled frame pairs or timestamps and require iterative inference loops, leading to relatively long runtimes.

\begin{table}[t]
\centering
\caption{Inference efficiency comparison evaluated on 16-frame clips.}
\label{tab:eff}
\resizebox{0.9\linewidth}{!}{
\begin{tabular}{lcccc}
\toprule
Method & FLOPs$\!$~(T) & Params$\!$~(M) & Memory$\!$~(MB) & Runtime$\!$~(s) \\
\midrule
St4RTrack\cite{st4rtrack} & 629.52 & 575.4 & \textbf{3273} & 20.59 \\
VDPM\cite{vdpm} & 333.72 & 1660.0 & 15749 & 11.70 \\
TraceAnything\cite{traceanything} & \textbf{8.79} & 674.5 & \underline{12613} & \textbf{1.05} \\
\midrule
Ours w/o sparse design & 15.25 & \textbf{568.7} & 23727 & 4.10 \\
Ours & 	\underline{12.07} & 	\textbf{568.7} & 21999 & \underline{2.15} \\
\bottomrule
\end{tabular}
}
\end{table}
\begin{table}[!htb]
\centering
\caption{Ablation study of SSA components. SelfAttn denotes self attention among temporally expanded dynamic tokens. Due to limited GPU memory, it is implemented sequentially as temporal attention followed by spatial attention.}
\label{tab:omnix_ablation_study}
\resizebox{\linewidth}{!}{ 
\begin{tabular}{cccc ccc} 
\toprule
\multicolumn{4}{c}{\textbf{Components}} & \multicolumn{3}{c}{\textbf{APD3D($\uparrow$)}} \\
\cmidrule(lr){1-4} \cmidrule(lr){5-7}
Eq.~\eqref{eq:temporal embed}-embed. & AdaLN-embed. & SelfAttn & CrossAttn & Monocular & Disjoint & Hybrid \\
\midrule
 & \checkmark & \checkmark & \checkmark & 0.260 & 0.267 & 0.291 \\
\checkmark &  & \checkmark & \checkmark & \underline{0.293} & \textbf{0.298} & \textbf{0.324} \\
\checkmark &  & \checkmark &  & 0.171 & 0.197 & 0.205 \\
\checkmark &  &  & \checkmark & \textbf{0.297} & \textbf{0.298} & \textbf{0.324} \\
\bottomrule
\end{tabular}
}
\end{table}

\subsection{Ablation Study}\label{sec:ablation}
\begin{figure}[htbp]
    \centering 
    \includegraphics[width=\textwidth]{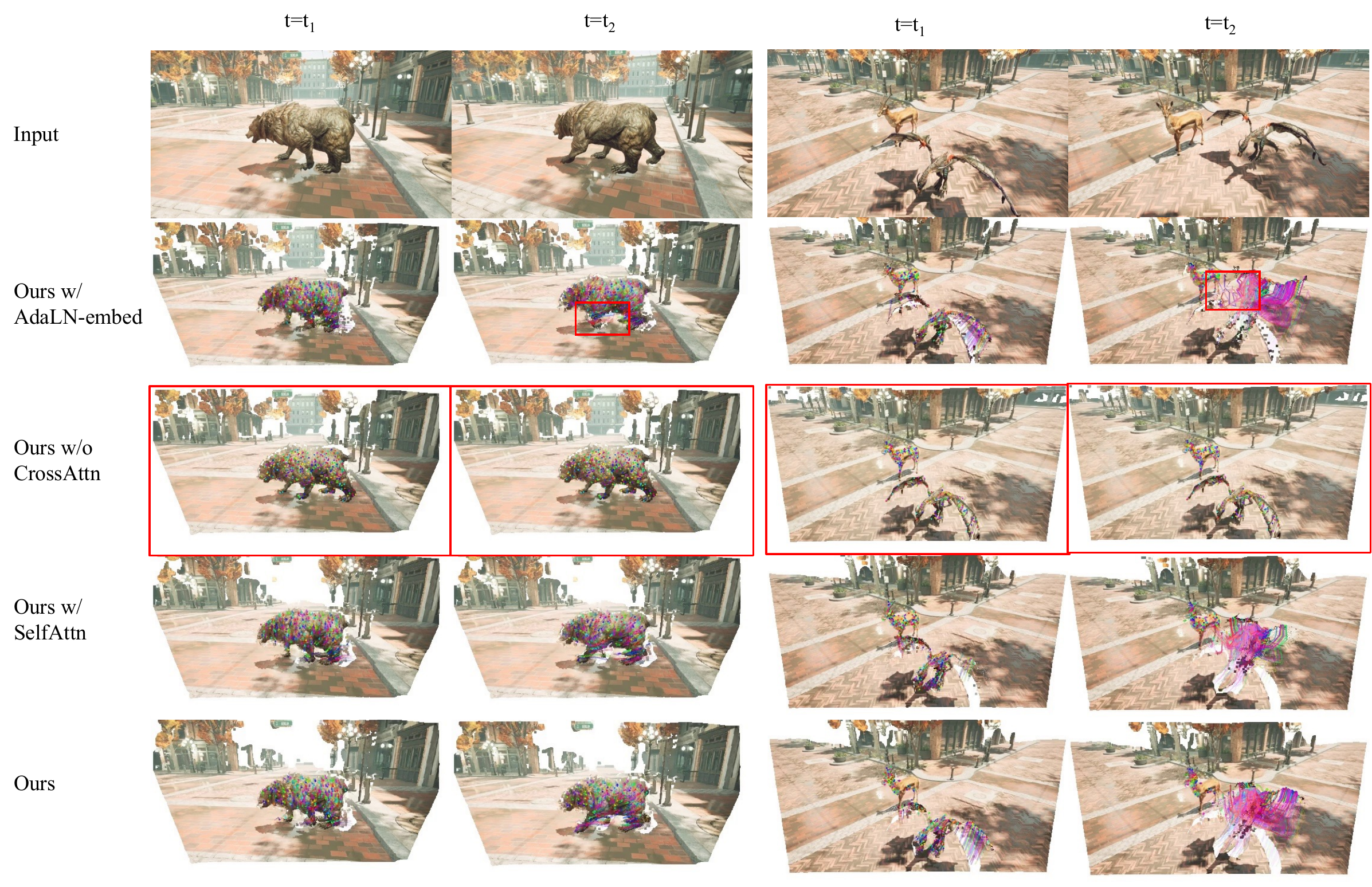} 
    \caption{Qualitative comparison for the ablation study of SSA components.} 
    \label{fig:ablate_SSA} 
\end{figure}

All ablation models are trained exclusively on our custom UE dataset. We primarily report the foreground APD3D metric since other metrics exhibit marginal variance across different settings, and our primary focus lies in capturing foreground motion dynamics. \cref{tab:omnix_ablation_study} summarizes the ablation of core components within the proposed SSA module.
1) Temporal Embedding Strategy: The proposed embedding strategy in Eq.~\eqref{eq:temporal embed} outperforms the widely adopted AdaLN~\cite{dit} mechanism.
2) Redundancy of Self Attention: We apply self attention to temporally expanded dynamic tokens using a factorized approach, \ie, conducting attention solely along the temporal dimension and then followed by the spatial dimension. Surprisingly, these self-attention layers in the SSA module yield negligible performance gains. This indicates that the essential spatiotemporal reasoning is already effectively captured by other components. Consequently, we discard these layers in our final model to improve computational efficiency.
3) Crucial Role of Cross Attention: Cross attention is vital for motion prediction. Removing it (see last two rows) leads to a drastic performance drop, fundamentally causing the model to fail in capturing foreground dynamics, \ie, mistakenly predicting static foregrounds.
Qualitative comparisons for the SSA ablation study are visualized in~\cref{fig:ablate_SSA}. Specifically, our model with AdaLN exhibits restricted motion in the bear's foot and erratic movement in the dragon's wings. In contrast, "Ours w/ SelfAttn" yields results similar to "Ours", whereas "Ours w/o CrossAttn" produces nearly static predictions. These results further support our quantitative analysis. 
\cref{tab:top_p_ablation} presents the ablation study on the hyper-parameter of top-$\rho\%$, and we use $\rho=20$ for efficiency during training and inference. \cref{tab:data_scaling_ablation} further shows the effect of data scaling, indicating that larger training datasets lead to better accuracy.

\begin{table}[t]
\centering
\begin{minipage}{0.48\linewidth}
    \centering
    \caption{Top-$\rho\%$ hyperparameter study on 4-frame monocular videos.}
    \label{tab:top_p_ablation}
    \footnotesize
    \begin{tabular}{l c c c}
        \toprule
        Top-$\rho\%$ & 10 & 20 & 30 \\
        \midrule
        APD3D$\!$ $\uparrow$ & 0.3194 & 0.3356 & \textbf{0.3454} \\
        \bottomrule
    \end{tabular}
\end{minipage}
\hfill
\begin{minipage}{0.48\linewidth}
    \centering
    \caption{Data scaling study on 16-frame monocular videos.}
    \label{tab:data_scaling_ablation}
    \footnotesize
    \begin{tabular}{l c c c}
        \toprule
        Training Size & 2.8k & 6.9k & 13.9k \\
        \midrule
        APD3D$\!$ $\uparrow$ & 0.182 & 0.210 & \textbf{0.243} \\
        \bottomrule
    \end{tabular}
\end{minipage}
\vspace{-4mm}
\end{table}

\section{Conclusion}
In this paper, we present \textbf{OmniX} , a novel feed-forward framework for dense 4D dynamic scene reconstruction. To overcome the limitation of previous methods in handling large camera motions, we explicitly disentangle dynamic foreground motions from static background structures. 
To address the scarcity of 4D training data, we developed a UE5-based data engine, constructing a large-scale dataset of 80K scenes and 1.28M multi-view videos with dense geometric and trajectory annotations. Extensive experiments demonstrate that OmniX sets a new state-of-the-art in dense 3D point trajectory prediction and 3D point tracking, and meanwhile achieves competitive results on video depth and camera pose estimation. We believe our framework and the newly introduced data engine will serve as a strong foundation for future research in dynamic scene understanding.

\subsubsection{Acknowledgments.} This work was supported in part by the Beijing Natural Science Foundation (Grant No. L223003), the Natural Science Foundation of China (Grant No. 62422317, U22B2056, 62192782, 62532015, 62403462, U25A20480), and 2025 Tencent AI Lab Rhino-Bird Focused Research Program.
%
%
\bibliographystyle{splncs04}
\bibliography{main}
\end{document}